\documentclass[runningheads]{llncs}
\usepackage{times}
\usepackage{soul}
\usepackage{url}
\usepackage[hidelinks]{hyperref}
\usepackage[utf8]{inputenc}
\usepackage[small]{caption}
\usepackage{graphicx}
\usepackage{amsmath}
\usepackage{booktabs}
\usepackage{enumitem}
\usepackage{graphicx}
\usepackage[ruled,vlined,linesnumbered]{algorithm2e}
\begin{document}

\title{Multi-agent Path Finding with Continuous Time Viewed Through Satisfiability Modulo Theories (SMT)}

\author{Pavel Surynek\orcidID{0000-0001-7200-0542}}
%
\authorrunning{P. Surynek}

%
\institute{
Faculty of Information Technology\\
Czech Technical University in Prague\\
Th\'{a}kurova 9, 160 00 Praha 6, Czech Republic\\
\email{pavel.surynek@fit.cvut.cz}
}
\maketitle              


\begin{abstract}
This paper addresses a variant of multi-agent path finding (MAPF) in continuous space and time. We present a new solving approach based on satisfiability modulo theories (SMT) to obtain makespan optimal solutions. The standard MAPF is a task of navigating agents in an undirected graph from given starting vertices to given goal vertices so that agents do not collide with each other in vertices of the graph. In the continuous version (MAPF$^\mathcal{R}$) agents move in an $n$-dimensional Euclidean space along straight lines that interconnect predefined positions. For simplicity, we work with circular omni-directional agents having constant velocities in the 2D plane. As agents can have different sizes and move smoothly along lines, a non-colliding movement along certain lines with small agents can result in a collision if the same movement is performed with larger agents. Our SMT-based approach for MAPF$^\mathcal{R}$ called SMT-CBS$^\mathcal{R}$ reformulates the Conflict-based Search (CBS) algorithm in terms of SMT concepts. We suggest lazy generation of decision variables and constraints. Each time a new conflict is discovered, the underlying encoding is extended with new variables and constraints to eliminate the conflict. We compared SMT-CBS$^\mathcal{R}$ and adaptations of CBS for the continuous variant of MAPF experimentally. 

\keywords{multi-agent path finding (MAPF), satisfiability modulo theory (SMT), continuous time, continuous space, makespan optimal solutions}
\end{abstract}

\section{Introduction and Background}

In {\em multi-agent path finding} (MAPF) \cite{DBLP:conf/focs/KornhauserMS84,DBLP:conf/aiide/Silver05,DBLP:journals/jair/Ryan08,DBLP:conf/icra/Surynek09,DBLP:journals/jair/WangB11,DBLP:journals/ai/SharonSGF13,SharonSFS15} the task is to navigate agents from given starting positions to given individual goals. The problem takes place in undirected graph $G=(V,E)$ where agents from set $A=\{a_1,a_2,...,a_k\}$ are placed in vertices with at most one agent per vertex. The initial configuration of agents can be written as $\alpha_0: A \rightarrow V$ and similarly the goal configuration as $\alpha_+: A \rightarrow V$.
In the standard MAPF, movements are instantaneous and are possible into vacant neighbors assuming no other agent is entering the same target vertex. We usually denote the configuration of agents at discrete time step $t$ as $\alpha_t: A \rightarrow V$. Non-conflicting movements transform configuration  $\alpha_t$ instantaneously into  $\alpha_{t+1}$ so we do not consider what happens between $t$ and $t+1$.

In order to reflect various aspects of real-life applications variants of MAPF have been introduced such as those considering {\em kinematic constraints} \cite{DBLP:conf/ijcai/HonigK00XAK17}, {\em large agents} \cite{LargeAAAI2019}, or {\em deadlines} \cite{DBLP:conf/ijcai/0001WFLKK18} - see \cite{DBLP:journals/corr/0001KA0HKUXTS17} for more variants. Particularly in this work we are dealing with an extension of MAPF introduced only recently \cite{DBLP:journals/corr/abs-1901-05506} that considers continuous time and continuous movements of agents between predefined positions (placed arbitrarily in the continuous space) denoted MAPF$^\mathcal{R}$. The contribution of this paper consists in showing how to apply satisfiability modulo theory (SMT) reasoning \cite{DBLP:conf/cp/Nieuwenhuis10,DBLP:journals/constraints/BofillPSV12} in MAPF$^\mathcal{R}$ solving.

\subsection{Related Work and Contributions}

Our SMT-based approach focuses on {\em makespan optimal} MAPF solving and builds on top of the Conflict-based Search (CBS) algorithm \cite{DBLP:conf/aaai/SharonSFS12,SharonSFS15}. Makespan optimal solutions minimize overall time needed to relocate all agents into their goals.

CBS tries to solve MAPF lazily by adding constraints on demand. It starts with an incomplete set of constraints which is iteratively refined with new conflict elimination constraints after conflicts are found in solutions for the incomplete set of constraints. Since conflict elimination constraints are disjuntive (they forbid occurrence of one or the other agent in a vertex at a time) the refinement is carried out as branching in the search process.

CBS can be adapted for MAPF$^\mathcal{R}$ by implementing conflict detection in continuous time and space \cite{DBLP:journals/corr/abs-1901-05506} while the high-level framework of the CBS algorithm remains the same. In the SMT-based approach we are trying to build an {\em incomplete} propositional model so that if a given MAPF$^\mathcal{R}$ $\Sigma^\mathcal{R}$ has a solution of a specified makespan then the model is solvable. This is similar to previous SAT-based \cite{Biere:2009:HSV:1550723} MAPF solving \cite{DBLP:conf/pricai/Surynek12,SurynekFSB16} where a {\em complete} propositional model has been constructed (that is, the given MAPF has a solution of a specified makespan if and only is the model is solvable). The propositional model in the SMT-based approach in constructed through conflict elimination refinements as done in CBS. The incompleteness of the model is inherited from CBS that adds constraints lazily (in contrast to this SAT-based methods like MDD-SAT \cite{SurynekFSB16} add all constraints eagerly). We call our new algorithm SMT-CBS$^\mathcal{R}$. The major difference of SMT-CBS$^\mathcal{R}$ from CBS is that instead of branching the search we only add a disjunctive constraint to eliminate the conflict in SMT-CBS$^\mathcal{R}$. Hence, SMT-CBS$^\mathcal{R}$ does not branch at all at the high-level (the model is incrementally refined at the high-level).

Similarly as in the SAT-based MAPF solving we use decision propositional variables indexed by {\em agent} $a$, {\em vertex} $v$, and {\em time} $t$ with natural meaning that if $TRUE$ agent $a$ appears in $v$ at time $t$. However the major technical difficulty with the continuous version of MAPF is that we do not know all necessary decision variables in advance. After a conflict is discovered we may need new decision variables to avoid that conflict. For this reason we introduce decision variable generation algorithm.

The paper is organized as follows: we first introduce MAPF$^\mathcal{R}$ formally. Then we recall a variant of CBS for MAPF$^\mathcal{R}$. Details of the novel SMT-based solving algorithm SMT-CBS$^\mathcal{R}$ follow. Finally an experimental evaluation of SMT-CBS$^\mathcal{R}$ against continuous version of CBS is shown. We also show a brief comparison with the standard MAPF.

\subsection{MAPF with Continuous Time}
We follow the definition of MAPF with continuous time denoted MAPF$^\mathcal{R}$ from \cite{DBLP:journals/corr/abs-1901-05506}.

\begin{definition} ({\bf MAPF$^\mathcal{R}$}) Multi-agent path finding with continuous time (MAPF$^\mathcal{R}$) is a 5-tuple $\Sigma^\mathcal{R}=(G=(V,E), A, \alpha_0, \alpha_+, \rho)$ where $G$, $A$, $\alpha_0$, $\alpha_+$ are from the standard MAPF and $\rho$ determines continuous extensions as follows:
\begin{itemize}
	\item $\rho.x(v), \rho.y(v)$ for $v \in V$ represent the position of vertex $v$ in the 2D plane; to simplify notation we will use $x_v$ for $\rho.x(v)$ and $y_v$ for $\rho.x(v)$
	\item $\rho.velocity(a)$ for $a \in A$ determines constant velocity of agent $a$; simple notation $v_a = \rho.velocity(a)$
	\item $\rho.diameter(a)$ for $a \in A$ determines diameter of agent $a$; we assume that agents are circular discs with omni-directional ability of movements; simple notation $d_a = \rho.diameter(a)$
\end{itemize}
\end{definition}

Naturally we can define the distance between a pair of vertices $u$, $v$ with $\{u,v\} \in E$ as $dist(u,v) = \sqrt{(x_v - x_u)^2+(y_v - y_u)^2}$.
Next we assume that agents have constant speed, that is, they instantly accelerate to $v_a$ from an idle state. The major difference from the standard MAPF where agents move instantly between vertices is that in MAPF$^\mathcal{R}$ continuous movement of an agent between a pair of vertices (positions) along the straight line interconnecting them takes place. Hence we need to be aware of the presence of agents at some point in the plane on the lines interconnecting vertices at any time.

Collisions may occur between agents due to their size which is another difference from the standard MAPF. In contrast to the standard MAPF, collisions in MAPF$^\mathcal{R}$ may occur not only in vertices but also on edges (on lines interconnecting vertices). If for example two edges are too close to each other and simultaneously traversed by large agents then such a condition may result in a collision.

We can further extend the set of continuous properties by introducing the direction of agents and the need to rotate agents towards the target vertex before they start to move towards the target (agents are no more omni-directional). The speed of rotation in such a case starts to play a role. Another interesting extension are agents of various shapes \cite{LargeAAAI2019}.
For simplicity we elaborate our solving concepts for the basic continuous extension of MAPF only. We however note that all developed concepts can be adapted for MAPF with more continuous extensions like directional agents, etc.

A solution to given MAPF$^\mathcal{R}$ $\Sigma^\mathcal{R}$ is a collection of temporal plans for individual agents $\pi = [\pi(a_1),\pi(a_2),...,\pi(a_k)]$ that are mutually collision-free. A temporal plan for agent $a \in A$ is a sequence $\pi(a) = [((\alpha_0(a),\alpha_1(a)),[t_0(a),t_1(a)));$ $((\alpha_1(a),\alpha_2(a)),$ $[t_1(a),t_2(a)));$ ...;$((\alpha_{m(a)-1},\alpha_{m(a)}(a)),(t_{m(a)-1},t_{m(a)}))]$ where $m(a)$ is the length of individual temporal plan and each pair $(\alpha_i(a),\alpha_{i+1}(a)), [t_i(a),t_{i+1}(a)))$ in the sequence corresponds to traversal event between vertices $\alpha_i(a)$ and $\alpha_{i+1}(a)$ starting at time $t_i(a)$ and finished at $t_{i+1}(a)$ (excluding).

It holds that $t_i(a) < t_{i+1}(a)$ for $i = 0,1,...,m(a)-1$. Moreover consecutive vertices must correspond to edge traversals or waiting actions, that is: $\{\alpha_i(a),$ $\alpha_{i+1}(a)\} \in E$ or $\alpha_i(a) = \alpha_{i+1}(a)$; and times must reflect the speed of agents for non-wait actions: $\alpha_i(a) \neq \alpha_{i+1}(a) \Rightarrow t_{i+1}(a) - t_i(a) = dist(\alpha_i(a), \alpha_{i+1}(a)) / v_a)$. In addition to this, agents must not collide with each other which is formally described in the following definition: 

\begin{definition} ({\bf collision}) A {\em collision} between individual temporal plans $\pi(a) = [((\alpha_i(a),$ $\alpha_{i+1}(a)),$ $[t_i(a),t_{i+1}(a)))]_{i=0}^{m(a)}$ and $\pi(b) = [((\alpha_i(b),\alpha_{i+1}(a)),[t_i(b),t_{i+1}(b)))]_{i=0}^{m(b)}$ occurs if the following condition holds:

\begin{itemize}
 	\item $\exists i \in \{0,1,...,m(a)\}$ and $\exists j \in \{0,1,...,m(b)\}$ such that:
 	
 	\begin{itemize}
 	\item $dist([x_{\alpha_i(a)},$ $y_{\alpha_i(a)};$ $x_{\alpha_{i+1}(a)},$ $y_{\alpha_{i+1}(a)}];$ $[x_{\alpha_j(b)},$ $y_{\alpha_j(b)};$ $x_{\alpha_{j+1}(b)},$ $y_{\alpha_{j+1}(b)}])$ $< d_a + d_b$
 	\item $[t_i(a),t_{i+1}(a)) \cap$ $[t_j(b),$ $t_{j+1}(b)) \neq \emptyset $
 	\end{itemize}
	(a vertex or an edge collision - two agents simultaneously occupy the same vertex or the same edge or traverse edges that are too close to each other)
\end{itemize}
\end{definition}

The distance between two lines $P$ and $Q$ given by their endpoint coordinates $P=[x_1,y_1;x_2,y_2]$ and $Q=[x'_1,y'_1;x'_2,y'_2]$ denoted $dist([x_1,y_1;x_2,y_2];[x'_1,y'_1;x'_2,y'_2])$ is defined as the minimum distance between any pair of points $p \in P$ and $q \in Q$: $min\{dist(p,q)\;|\;p \in P \wedge q \in Q\}$. The definition covers degenerate cases where a line collapses into a single point. In such a case the definition of $dist$ normally works as the distance between points and between a point and a line.

The definition among other types of collisions covers also a case when an agent waits in vertex $v$ and another agent passes through a line that is too close to $v$. We note that situations classified as collisions according to the above definition may not always result in actual collisions where agents' bodies overlap; the definition is overcautious in this sense. Alternatively we can use more precise definition of collisions that reports collisions if and only if an actual overlap of agents' bodies occurs. This however requires more complex equations or simulations.

\begin{figure}[h]
    \centering
    \vspace{-0.2cm}
    \includegraphics[trim={2.0cm 22.0cm 2.0cm 3cm},clip,width=1.0\textwidth]{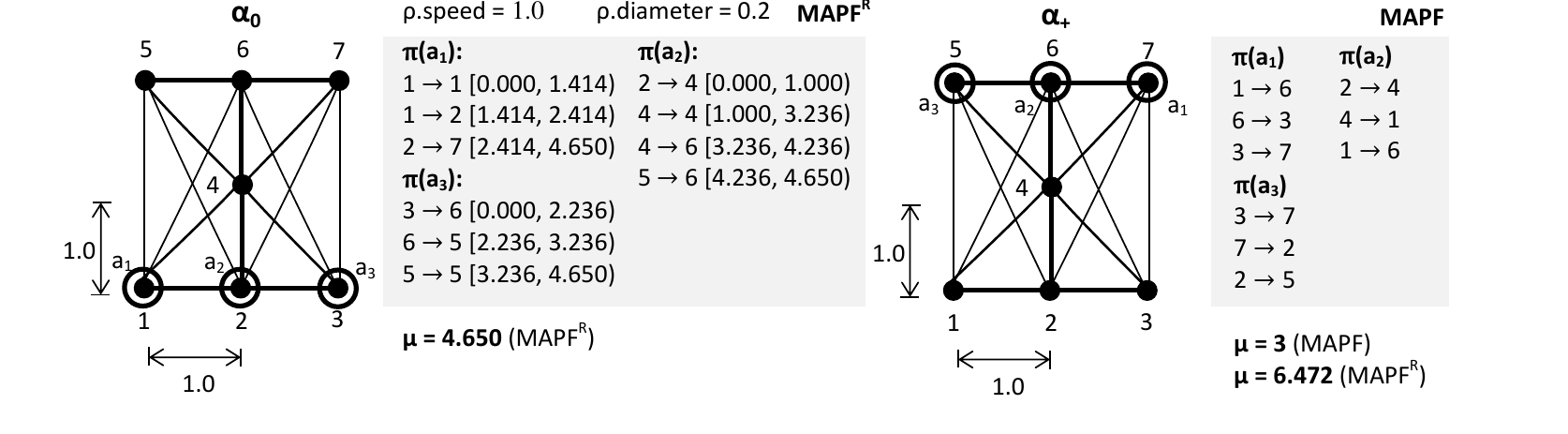}
    \vspace{-0.8cm}
    \caption{An example of MAPF$^\mathcal{R}$ instance on a $[3,1,3]-$graph with three agents and its makespan optimal solution (an optimal solution of the corresponding standard MAPF is shown too).}
    \vspace{-0.2cm}
    \label{fig-MAPF-R}
\end{figure}

The duration of individual temporal plan $\pi(a)$ is called an individual makespan; denoted $\mu(\pi(a)) = t_{m(a)}$. The overall makespan of MAPF$^\mathcal{R}$ solution $\pi = [\pi(a_1),\pi(a_2),$ ...$,\pi(a_k)]$ is defined as max$_{i=1}^k(\mu(\pi(a_i)))$. In this work we focus on finding makespan optimal solutions. An example of MAPF$^\mathcal{R}$ and makespan optimal solution is shown in Figure \ref{fig-MAPF-R}. We note that the standard makespan optimal solution yields makespan suboptimal solution when interpreted as MAPF$^\mathcal{R}$.

Through the straightforward reduction of MAPF to MAPF$^\mathcal{R}$ it can be observed that finding a makespan optimal solution with continuous time is an NP-hard problem \cite{DBLP:journals/jsc/RatnerW90,DBLP:conf/aaai/Surynek10,DBLP:journals/corr/YuL15c}.

\section{Solving MAPF with Continuous Time}

We will describe here how to find optimal solution of MAPF$^\mathcal{R}$ using the {\em conflict-based search} (CBS) \cite{SharonSFS15}. CBS uses the idea of resolving conflicts lazily; that is, a solution of MAPF instance is not searched against the complete set of movement constraints. Instead of forbidding all possible collisions between agents we start with initially empty set of collision forbidding constraints that gradually grows as new conflicts appear. CBS originally developed for MAPF can be easily modified for MAPF$^\mathcal{R}$ as shown in \cite{DBLP:journals/corr/abs-1901-05506}: let us call the modification CBS$^\mathcal{R}$.

\subsection{Conflict-based Search}

CBS$^\mathcal{R}$ is shown using pseudo-code in Algorithm \ref{alg-CBS-R}. The high-level of CBS$^\mathcal{R}$ searches a {\em constraint tree} (CT) using a priority queue (ordered according to the makespan or other cumulative cost) in the breadth first manner. CT is a binary tree where each node $N$ contains a set of collision avoidance constraints $N.constraints$ - a set of triples $(a_i,\{u,v\},[t_0,t_+))$ forbidding occurrence of agent $a_i$ in edge $\{u,v\}$ (or in vertex $u$ if $u = v$) at any time between $[t_0,t_+)$, a solution $N.\pi$ - a set of $k$ individual temporal plans, and the makespan $N.\mu$ of the current solution.

\begin{algorithm}[h]
\begin{footnotesize}
\SetKwBlock{NRICL}{CBS$^\mathcal{R}$ ($\Sigma^\mathcal{R}=(G=(V,E), A, \alpha_0, \alpha_+, \rho)$)}{end} \NRICL{
    $R.constraints \gets \emptyset$ \\
    $R.\pi \gets$ $\{$shortest temporal plan from $\alpha_0(a_i)$ to $\alpha_+(a_i)\;|\;i = 1,2,...,k\}$\\
    $R.\mu \gets \max_{i=1}^k{\mu(N.\pi(a_i))}$ \\
    $\textsc{Open} \gets \emptyset$ \\     
    insert $R$ into $\textsc{Open}$ \\
    \While {$\textsc{Open} \neq \emptyset$} {
        $N \gets$ min$_{\mu}(\textsc{Open}$)\\
        remove-Min$_{\mu}(\textsc{Open}$)\\
        $collisions \gets$ validate-Plans($N.\pi$)\\
        \If {$collisions = \emptyset$}{
            \Return $N.\pi$\\
        }
        let $(a_i,\{u,v\},[t_0,t_+)) \times (a_j,\{u',v'\},[t'_0,t'_+)) \in collisions$\\
        $[\tau_0, \tau_+) \gets [t_0,t_+) \cap [t'_0,t'_+)$ \\
        
        \For {each $(a,\{w,z\}) \in \{(a_i,\{u,v\}),(a_j,\{u',v'\})\}$}{
       	$N'.constraints \gets N.constraints \cup \{(a,\{w,z\},[\tau_0,\tau_+))\}$\\
        	$N'.\pi \gets N.\pi$\\
        	update($a$, $N'.\pi$, $N'.conflicts$)\\
		$N'.\mu \gets \sum_{i=1}^k\mu{(N'.\pi(a_i))}$\\
		insert $N'$ into $\textsc{Open}$ \\
        }
     }
} \caption{Basic CBS$^\mathcal{R}$ algorithm for solving MAPF with continuous time} \label{alg-CBS-R}
\end{footnotesize}
\end{algorithm}

The low-level process in CBS$^\mathcal{R}$ associated with node $N$ searches temporal plan for individual agent with respect to set of constraints $N.constraints$. For given agent $a_i$, this is the standard single source shortest path search from $\alpha_0(a_i)$ to $\alpha_+(a_i)$ that at time $t$ must avoid a set of edges (vertices) $\{\{u,v\} \in E\;|\;(a_i,\{u,v\},[t_0,t_+)) \in N.constraints \wedge t \in [t_0,t_+)\}$.

CBS$^\mathcal{R}$ stores nodes of CT into priority queue $\textsc{Open}$ sorted according to the ascending makespan. At each step CBS takes node $N$ with the lowest makespan from $\textsc{Open}$ and checks if $N.\pi$ represent non-colliding temporal plans. If there is no collision, the algorithms returns valid MAPF$^\mathcal{R}$ solution $N.\pi$. Otherwise the search branches by creating a new pair of nodes in CT - successors of $N$. Assume that a collision occurred between agents $a_i$ and $a_j$ when $a_i$ traversed $\{u,v\}$ during $[t_0,t_+)$ and $a_j$ traversed $\{u',v'\}$ during $[t'_0,t'_+)$. This collision can be avoided if either agent $a_i$ or agent $a_j$ does not occupy $\{u,v\}$ or $\{u',v'\}$ respectively during $[t_0,t_+) \cap [t'_0,t'_+) = [\tau_0, \tau_+)$. These two options correspond to new successor nodes of $N$: $N_1$ and $N_2$ that inherit set of conflicts from $N$ as follows: $N_1.conflicts = N.conflicts$ $\cup \{(a_i,\{u,v\},[\tau_0, \tau_+))\}$ and $N_2.conflicts = N.conflicts$ $\cup \{(a_j,\{u',v'\},[\tau_0, \tau_+))\}$. $N_1.\pi$ and $N_1.\pi$ inherit plans from $N.\pi$ except those for agent $a_i$ and $a_j$ respectively that are recalculated with respect to the new sets of conflicts. After this $N_1$ and $N_2$ are inserted into $\textsc{Open}$.

Definition of collisions comes as a parameter to the algorithm though the implementation of validate-Plans procedure. We can switch to the less cautious definition of collisions that reports a collision after agents actually overlap their bodies. This can be done through changing the validate-Plans procedure while the rest of the algorithm remains the same.

\subsection{A Satisfiability Modulo Theory (SMT) Approach}

A close look at CBS reveals that it operates similarly as problem solving in {\em satisfiability modulo theories} (SMT) \cite{DBLP:journals/constraints/BofillPSV12,DBLP:conf/cp/Nieuwenhuis10}. The basic application of SMT divides satisfiability problem in some complex theory $T$ into an abstract propositional part that keeps the Boolean structure of the decision problem and a simplified decision procedure $DECIDE_T$ that decides fragment of $T$ restricted on {\em conjunctive formulae}. A general $T$-formula $\Gamma$ is transformed to a {\em propositional skeleton} by replacing atoms with propositional variables. The standard SAT-solving procedure then decides what variables should be assigned $TRUE$ in order to satisfy the skeleton - these variables tells what atoms hold in $\Gamma$. $DECIDE_T$ then checks if the conjunction of atoms assigned $TRUE$ is valid with respect to axioms of $T$. If so then satisfying assignment is returned and we are finished. Otherwise a conflict from $DECIDE_T$ (often called a lemma) is reported back to the SAT solver and the skeleton is extended with new constraints resolving the conflict. More generally not only new constraints are added to resolve a conflict but also new variables i.e. atoms can be added to $\Gamma$.

The above observation led us to the idea to rephrase CBS$^\mathcal{R}$ in terms of SMT. A plan validation procedure will act as $DECIDE_T$ and will report back a set of conflicts found in the current solution. Hence axioms of $T$ will be represented by the movement rules of MAPF$^\mathcal{R}$. The propositional part working with the skeleton will be taken from existing propositional encodings of the standard MAPF such as the MDD-SAT \cite{SurynekFSB16} provided that constraints forbidding conflicts between agents will be omitted (at the beginning). In other words, we only preserve constraints ensuring that propositional assignments form proper paths for agents but each agent is treated as if it is alone in the instance.

\subsection{Decision Variable Generation}

MDD-SAT introduces decision variables $\mathcal{X}_{v}^{t}(a_i)$  and $\mathcal{E}_{u,v}^{t}(a_i)$ for discrete time-steps $t \in \{0,1,2, ...\}$ describing occurrence of agent $a_i$ in $v$ or the traversal of edge $\{u,v\}$ by $a_i$ at time-step $t$. We refer the reader to \cite{SurynekFSB16} for the details of how to encode constraints of top of these variables. As an example we show here a constraint stating that if robot $a_i$ appears in vertex $u$ at time step $t$ then it has to leave through exactly one edge connected to $u$ or wait in $u$.

\begin{equation}
   {  \mathcal{X}_u^t(a_i) \Rightarrow \bigvee_{v\;|\;\{u,v\} \in E}{\mathcal{E}^t_{u,v}(a_i)} \vee \mathcal{E}^t_{u,u}(a_i),
   }
   \label{eq-1}
\end{equation}
\begin{equation}
   {  \sum_{v\:|\:\{u,v\} \in E }{\mathcal{E}_{u,v}^t{(a_i)} + \mathcal{E}^t_{u,u}(a_i) \leq 1}
   }
   \label{eq-2}
\end{equation}

Vertex collisions expressed for example by the following constraint are omitted. The constraint say that in vertex $v$ and time step $t$ there is at most one agent.

\begin{equation}
    {\sum_{a_i \in A \:|\:v \in V}{\mathcal{X}^t_v(a_i)} \leq 1
    }
    \label{eq-3}
\end{equation}

A significant difficulty in MAPF$^\mathcal{R}$ is that we need decision variables with respect to continuous time. Fortunately we do not need a variable for any possible time but only for important moments.

If for example the duration of a conflict in neighbor $v$ of $u$ is $[t_0,t_+)$ and agent $a_i$ residing in $u$ at $t \geq t_0$ wants to enter $v$ then the earliest time $a_i$ can do so is $t_+$ since before it would conflict in $v$ (according to above definition of collisions). On the other hand if $a_i$ does not want to waste time, then waiting longer than $t_+$ is not desirable. Hence we only need to introduce decision variable $\mathcal{E}_{u,v}^{t_+}(a_i)$ to reflect the situation.

\begin{algorithm}[t]
\begin{footnotesize}
\SetKwBlock{NRICL}{generate-Decisions ($\Sigma^\mathcal{R}=(G=(V,E), A, \alpha_0, \alpha_+, \rho)$, $a_i$, $conflicts$, $\mu_{max}$)}{end} \NRICL{
    $\textsc{Var} \gets \emptyset$\\
    \For {each $a \in A$}{
    $\textsc{Open} \gets \emptyset$ \\   
    insert $(\alpha_0(a),0)$ into $\textsc{Open}$\\
    $\textsc{Var} \gets \textsc{Var} \cup \{  \mathcal{X}_{\alpha_0(a)}^{t_0}(a) \}$\\
    \While {$\textsc{Open} \neq \emptyset$} {
            $(u,t) \gets$ min$_{t}(\textsc{Open}$)\\
            remove-Min$_{t}(\textsc{Open}$) \\
            \If {$t \leq \mu_{max}$}{
	            \For {each $v$ such that $\{u,v\} \in E$}{
      		             $\Delta t \gets dist(u,v) / v_a$ \\
             		insert $(v,t+\Delta t)$ into $\textsc{Open}$\\             		
				$\textsc{Var} \gets \textsc{Var} \cup \{ \mathcal{E}_{u,v}^{t}(a), \mathcal{X}_{v}^{t+\Delta t}(a) \}$ \\
	            	}
      		      \For {each $v$ such that $\{u,v\} \in E \cup \{u,u\}$}{
            			\For {each $(a, \{u,v\}, [t_0,t_+)) \in conflicts$}{
            				\If {$t_+ > t$}{
             				insert $(u,t_+)$ into $\textsc{Open}$\\
						$\textsc{Var} \gets \textsc{Var} \cup \{  \mathcal{X}_{u}^{t_+}(a) \}$ \\
             			}
	            		}
      		      	}
		}
	  }
    }
    \Return $\textsc{Var}$\\
} \caption{Generation of decision variables in the SMT-based algorithm for MAPF$^\mathcal{R}$ solving} \label{alg-DEC-gen}
\end{footnotesize}
\end{algorithm}

Generally when having a set of conflicts we need to generate decision variables representing occurrence of agents in vertices and edges of the graph at important moments with respect to the set of conflicts. The process of decision variable generation is formally described as Algorithm \ref{alg-DEC-gen}. It performs breadth-first search (BFS) on G using two types of actions: {\em edge traversals} and {\em waiting}. The edge traversal is the standard operation from BFS. Waiting is performed for every relevant period of time with respect to the end-times in the set of conflicts of neighboring vertices.

As a result each conflict during variable generation through BFS is treated as both present and absent which in effect generates all possible important moments.

\subsection{Eliminating Branching in CBS by Disjunctive Refinements}

The SMT-based algorithm itself is divided into two procedures: SMT-CBS$^\mathcal{R}$ representing the main loop and SMT-CBS-Fixed$^\mathcal{R}$ solving the input MAPF$^\mathcal{R}$ for a fixed maximum makespan $\mu$. The major difference from the standard CBS is that there is no branching at the high level. The set of conflicts is iteratively collected during the entire execution of the algorithm whenever a collision is detected. 

Procedures {\em encode-Basic} and {\em augment-Basic} build formula $\mathcal{F}(\mu)$ over decision variables generated using the aforementioned procedure. The encoding is inspired by the MDD-SAT approach but ignores collisions between agents. That is, $\mathcal{F}(\mu)$ constitutes an {\em incomplete model} for a given input $\Sigma^\mathcal{R}$: $\Sigma^\mathcal{R}$ is solvable within makespan $\mu$ then $\mathcal{F}(\mu)$ is satisfiable.

Conflicts are resolved by adding disjunctive constraints (lines 13-15 in Algorithm \ref{alg-SMTCBS-low}). The collision is avoided in the same way as in the original CBS that is one of the colliding agent does not perform the action leading to the collision. Consider for example a collision on edges between $a_i$ and $a_j$: $a_i$ traversed $\{u,v\}$ during $[t_0,t_+)$ and $a_j$ traversed $\{u',v'\}$ during $[t'_0,t'_+)$. 

These two movements correspond to decision variables $\mathcal{E}_{u,v}^{t_0}(a_i)$ and $\mathcal{E}_{u',v'}^{t_0'}(a_j)$ hence elimination of the collision caused by these two movements can be expressed as following disjunction: $\neg \mathcal{E}_{u,v}^{t_0}(a_i) \vee \neg \mathcal{E}_{u',v'}^{t'_0}(a_j)$. The set of pairs of disjunctive conflicts is propagated across entire execution of the algorithm (line 16 in Algorithm \ref{alg-SMTCBS-low}).

\begin{algorithm}[h]
\begin{footnotesize}
\SetKwBlock{NRICL}{SMT-CBS$^\mathcal{R}$ ($\Sigma^\mathcal{R}=(G=(V,E), A, \alpha_0, \alpha_+, \rho)$)}{end} \NRICL{
    $conflicts \gets \emptyset$\\
    $\pi \gets$ $\{\pi^*(a_i)$ a shortest temporal plan from $\alpha_0(a_i)$ to $\alpha_+(a_i)\;|\;i = 1,2,...,k\}$ \\
    $\mu \gets \max_{i=1}^k{\mu(\pi(a_i))}$ \\
    \While {$TRUE$}{
         $(\pi,conflicts,\mu_{next}) \gets$ SMT-CBS-Fixed$^\mathcal{R}$($\Sigma^\mathcal{R}$, $conflicts$, $\mu$)\\
        \If {$\pi \neq$ UNSAT}{
        	\Return $\pi$\\
        }
        $\mu \gets \mu_{next}$\\
    }
}
 \caption{High-level of the SMT-based MAPF$^\mathcal{R}$ solving} \label{alg-SMTCBS-high}
\end{footnotesize}
\end{algorithm}

The iterative scheme for trying larger makespans follows the idea that for a fixed makespan $\mu_{max}$ we generate decision variables for all possible single movement continuations above $\mu_{max}$ (line 10 of Algorithm \ref{alg-DEC-gen}). The next makespan to try will then be obtained by taking the current makespan plus the smallest duration of the continuing movement (line 19 of Algorithm \ref{alg-SMTCBS-low}). The process ensures finding makespan optimal solution (proof omitted).

 \begin{algorithm}[h]
\begin{footnotesize}
\SetKwBlock{NRICL}{SMT-CBS-Fixed$^\mathcal{R}$($\Sigma^\mathcal{R}$, $conflicts$, $\mu$)}{end} \NRICL{
	    $\textsc{Var} \gets$ generate-Decisions($\Sigma^\mathcal{R}$, $conflicts$, $\mu$)\\
	    $\mathcal{F}(\mu) \gets$ encode-Basic$(\textsc{Var},\Sigma^\mathcal{R},\mu)$\\
	    \While {$TRUE$}{
	        $assignment \gets$ consult-SAT-Solver$(\mathcal{F}(\mu))$\\
	        \If {$assignment \neq UNSAT$}{
	            $\pi \gets$ extract-Solution$(assignment)$\\
	            $collisions \gets$ validate-Plans($\pi$)\\
                   \If {$collisions = \emptyset$}{
                      \Return $(\pi, UNDEF, conflicts)$\\
                   }
                   \For{each $(a_i,\{u,v\},[t_0,t_+)) \times (a_j,\{u',v;\},[t'_0,t'_+)) \in collisions$}{
                   	   $\mathcal{Y} \gets (u=v) \;\; ? \;\; \mathcal{X}_{u}^{t_0}(a_i) : \mathcal{E}_{u,v}^{t_0}(a_i)$\\
                   	   $\mathcal{Z} \gets (u'=v') \;\; ? \;\; \mathcal{X}_{u'}^{t'_0}(a_j) : \mathcal{E}_{u',v'}^{t'_0}(a_j)$\\
                   	   
                      $\mathcal{F}(\mu) \gets \mathcal{F}(\mu) \cup \{\neg \mathcal{Y} \vee \neg \mathcal{Z}$\}\\
                      $[\tau_0, \tau_+) \gets [t_0,t_+) \cap [t'_0,t'_+)$ \\
                      $conflicts \gets conflicts \cup \{(a_i,\{u,v\},[\tau_0,\tau_+)),(a_j,\{u',v'\},[\tau_0,\tau_+))\}$\\                   
                   }                  
	    		$\textsc{Var} \gets$ generate-Decisions($\Sigma^\mathcal{R}$, $conflicts$, $\mu$)\\
	    		$\mathcal{F}(\mu) \gets$ augment-Basic$(\textsc{Var},\mathcal{F}(\mu),\Sigma^\mathcal{R},\mu)$\\                   
               }
         }
	   $\mu_{next} \gets$ min$\{ t\;|\; \mathcal{X}_{u}^{t}(a_i) \in \textsc{Var} \wedge t > \mu)\}$ \\               
         \Return {(UNSAT, $conflicts$, $\mu_{next}$)}\\
}
\caption{Low-level of the SMT-based MAPF$^\mathcal{R}$ solving} \label{alg-SMTCBS-low}
\end{footnotesize}
\end{algorithm}

\section{Experimental Evaluation}

We implemented SMT-CBS$^\mathcal{R}$ in C++ to evaluate its performance. SMT-CBS$^\mathcal{R}$ was implemented on top of Glucose 4 SAT solver \cite{DBLP:conf/ijcai/AudemardS09} which ranks among the best SAT solvers according to recent SAT solver competitions \cite{DBLP:conf/aaai/BalyoHJ17}. Whenever possible the SAT solver was consulted in the incremental mode.

In addition to SMT-CBS$^\mathcal{R}$ we re-implemented in C++ CBS$^\mathcal{R}$, currently the only alternative solver for MAPF$^\mathcal{R}$ based on own dedicated search \cite{DBLP:journals/corr/abs-1901-05506}.

Our implementation of CBS$^\mathcal{R}$ used the standard heuristics to improve the performance such as the preference of resolving {\em cardinal conflicts} \cite{DBLP:conf/ijcai/BoyarskiFSSTBS15}. Without this heuristic, CBS$^\mathcal{R}$ usually exhibited poor performance. In the preliminary tests with SMT-CBS$^\mathcal{R}$, we initially tried to resolve against single cardinal conflict too but eventually it turned out to be more efficient to resolve against all discovered conflicts (the presented pseudo-code shows this variant). \footnote {All experiments were run on a system  with Ryzen 7 3.0 GHz, 16 GB RAM, under Ubuntu Linux 18.}


\subsection{Benchmarks and Setup}
SMT-CBS$^\mathcal{R}$ and CBS$^\mathcal{R}$ were tested on {\em layered graphs}. The layered graph of height $h$ denoted $[l_1,l_2,...,l_h]$-graph consists of $h$ layers of vertices placed horizontally above each other in the 2D plane (see Figure \ref{fig-MAPF-R} for $[3,1,3]$-graph). More precisely the $i$-th layer is placed horizontally at $y = i$. Layers are centered horizontally and the distance between consecutive points in the layer is $1.0$. Size of all agents was $0.2$ in diameter.

We measured runtime and the number of decisions/iterations to compare performance of SMT-CBS$^\mathcal{R}$ and CBS$^\mathcal{R}$. Small layered graphs consisting of 2 to 5 layers with up to 5 vertices per layer were used in tests. Three consecutive layers are fully interconnected by edges.


In all tests agents started in the $1$-st layer and finished in the last $h$-th layer. To obtain instances of various difficulties random permutations of agents in the starting and goal configurations were used (the $1$-st layer and $h$-th layer were fully occupied in the starting and goal configuration respectively). If for instance agents are ordered identically in the starting and goal configuration with $h \leq 3$, then the instance is relatively easy as it is sufficient that all agents move simultaneously straight into their goals.

Ten random instances were generated for each layered graph. The timeout for all tests has been set to 1 minute. Results from instances finished under this timeout were used to calculate average runtimes.

\subsection{Comparison of MAPF$^\mathcal{R}$ and MAPF Solving}

\begin{figure}[h]
    \centering
    \vspace{-0.4cm}
    \includegraphics[trim={2.0cm 13.0cm 5.0cm 2.0cm},clip,width=1.0\textwidth]{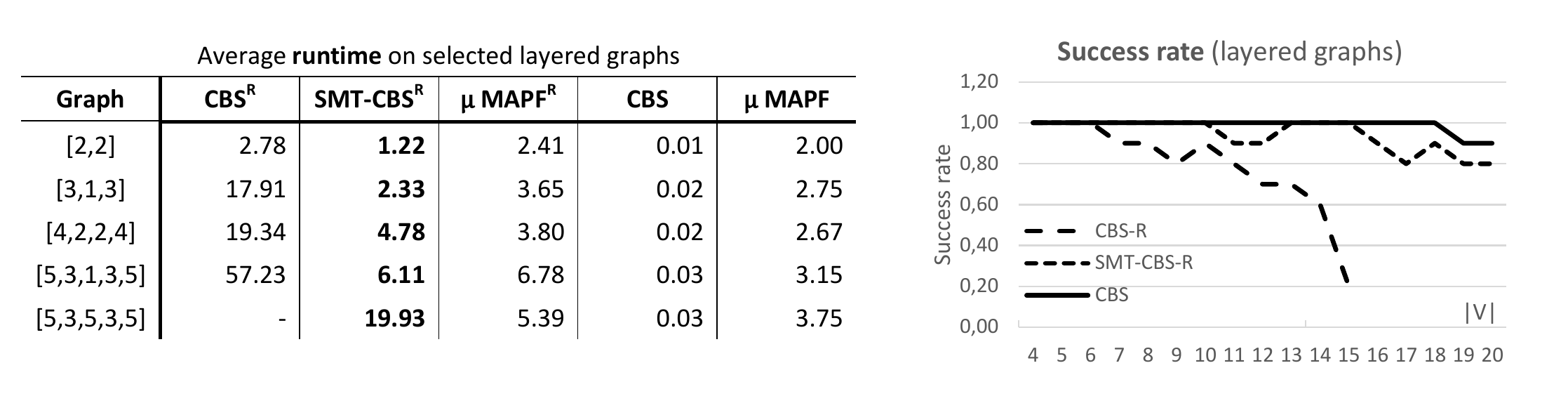}
    \vspace{-0.8cm}
    \caption{Comparison of CBS$^\mathcal{R}$ and SMT-CBS$^\mathcal{R}$ in terms of average runtime and success rate. The standard CBS on the corresponding standard MAPF is shown too (times are in seconds).}
    \vspace{-0.6cm}
    \label{expr_layered-runtime}
\end{figure}

Part of the results obtained in our experimentation is shown in Figure \ref{expr_layered-runtime}. The general observation from our runtime evaluation is that MAPF$^\mathcal{R}$ is significantly harder than the standard MAPF. When continuity is ignored, makespan optimal solutions consist of fewer steps. But due to regarding all edges to be unit in MAPF, the standard makespan optimal solutions yield significantly worse continuous makespan (this effect would be further manifested if we use longer edges).

SMT-CBS$^\mathcal{R}$ outperforms CBS$^\mathcal{R}$ on tested instances significantly. CBS$^\mathcal{R}$ reached the timeout many more times than SMT-CBS$^\mathcal{R}$. In the absolute runtimes, SMT-CBS$^\mathcal{R}$ is faster by factor of 2 to 10.

In terms of the number of decisions, SMT-CBS$^\mathcal{R}$ generates order of magnitudes fewer iterations than CBS$^\mathcal{R}$. This is however expected as SMT-CBS$^\mathcal{R}$ shrinks the entire search tree into a single branch in fact. We note that branching within the search space in case of SMT-CBS$^\mathcal{R}$ is deferred into the SAT solver where even more branches may appear.

\section{Discussion and Conclusion}

We suggested a novel approach for the makespan optimal solving of the multi-agent path finding problem with continuous time based on satisfiability modulo theories (SMT). Our approach is based on the idea of treating conflicts lazily as suggested in the CBS algorithm but instead of branching the search after encountering a conflict we refine the propositional model with the conflict elimination disjunctive constraint. The major obstacle in using SMT and propositional reasoning is that decision variables cannot be determined in advance straightforwardly in the continuous case. We hence suggested a novel decision variable generation approach that enumerates new decisions after discovering new conflicts. The propositional model is iteratively solved by the SAT solver. We call the novel algorithm SMT-CBS$^\mathcal{R}$.

We compared SMT-CBS$^\mathcal{R}$ to the only previous approach for MAPF$^\mathcal{R}$ that modifies the standard CBS algorithm \cite{DBLP:journals/corr/abs-1901-05506} and uses dedicated search. The outcome of our comparison is that SMT-CBS$^\mathcal{R}$ is significantly faster. We attribute the better runtime results of SMT-CBS$^\mathcal{R}$ to more efficient handling of disjunctive conflicts in the underlying SAT solver through {\em propagation}, {\em clause learning} and other mechanisms. In contrast to this CBS$^\mathcal{R}$ relies on its own dedicated search that is less advanced compared to SAT solvers.

Comparison with the standard MAPF version indicate that continuous reasoning is harder to solve but on the other hand provides more realistic solutions.

For the future work we assume extending the concept of SMT-based approach for MAPF$^\mathcal{R}$ with other cumulative cost functions other than the makespan such as the {\em sum-of-costs} \cite{ICTSJUR}.

\bibliographystyle{splncs04}
\bibliography{bibfile}

\end{document}